\documentclass{article}
\usepackage{ijcai22}

\usepackage{times}
\usepackage{soul}
\usepackage{url}
\usepackage[hidelinks]{hyperref}
\usepackage[utf8]{inputenc}
\usepackage[small]{caption}
\usepackage{graphicx}
\usepackage{amsmath}
\usepackage{amsthm}
\usepackage{booktabs}
\usepackage{algorithm}
\usepackage{hyperref}
\usepackage{multirow}
\usepackage{subfigure}
\usepackage{algorithm}
\usepackage{algpseudocode}

\newcommand{\SP}[1]{\ensuremath{^{\textrm{#1}}}}

\newcommand{\citet}[1]{\citeauthor{#1} \shortcite{#1}} \newcommand{\citep}{\cite} 

\title{CUP: Curriculum Learning based Prompt Tuning for Implicit \\ Event Argument Extraction}

\author{
Jiaju Lin$^1$
\and
Qin Chen$^1$\footnote{Corresponding author} \and
Jie Zhou$^{2}$\and
Jian Jin$^1$ \And
Liang He$^1$ \\
\affiliations
$^1$School of Computer Science and Technology, East China Normal University\\
$^2$School of Computer Science, Fudan University
\emails
jiaju\_lin@stu.ecnu.edu.cn,
\{qchen,jjin,lhe\}@cs.ecnu.edu.cn,
jie\_zhou@fudan.edu.cn
}

\begin{document}

\maketitle

\begin{abstract}

Implicit event argument extraction (EAE) aims to identify arguments that could scatter over the document. Most previous work focuses on learning the direct relations between arguments and the given trigger, while the implicit relations with long-range dependency are not well studied. Moreover, recent neural network based approaches rely on a large amount of labeled data for training, which is unavailable due to the high labelling cost. In this paper, we propose a \textbf{Cu}rriculum learning based \textbf{P}rompt tuning (CUP) approach, which resolves implicit EAE by four learning stages. The stages are defined according to the relations with the trigger node in a semantic graph, which well captures the long-range dependency between arguments and the trigger. In addition, we integrate a prompt-based encoder-decoder model to elicit related knowledge from pre-trained language models (PLMs) in each stage, where the prompt templates are adapted with the learning progress to enhance the reasoning for arguments. Experimental results on two well-known benchmark datasets show the great advantages of our proposed approach. In particular, we outperform the state-of-the-art models in both fully-supervised and low-data scenarios.

\end{abstract}

\section{Introduction}
\begin{figure}
    \centering
     \includegraphics[width=0.96\linewidth]{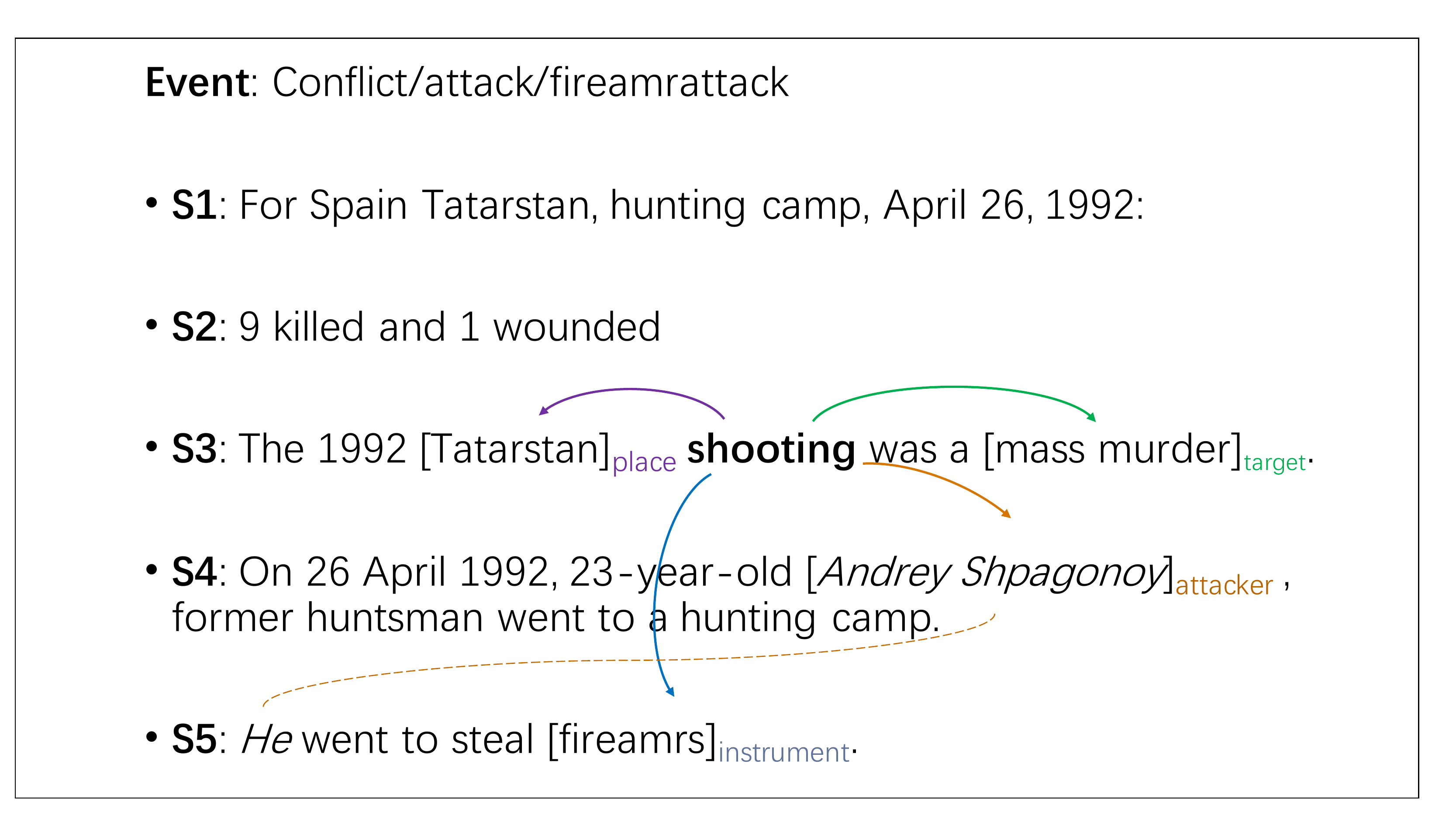}
    \caption{An example of implicit EAE from RAMS. Solid lines link the \textbf{trigger} and $\rm [argument]_{role}$. Dash lines link the $coreference\ entities$ across sentence boundaries. }
    \label{fig:intro}
\end{figure}

Event argument extraction (EAE) plays an important role in artificial intelligence, which has been widely used in global crisis monitoring and decision making \cite{zhan2019survey,mostafazadeh-davani-etal-2019-reporting,chen-etal-2020-joint-modeling}. Unlike traditional EAE that mainly focuses on extracting arguments within a single sentence given a trigger, implicit EAE focuses on document-level argument extraction that arguments may span over multiple sentences, which is more challenging \cite{ebner-etal-2020-RAMS}. As shown in Figure \ref{fig:intro}, the word \textit{`shooting'} triggers a \textit{Conflict/attack/fireamrattack} event. Implict EAE aims to extract arguments corresponding to their roles beyond the sentence level, namely \textit{Tatarstan (place)}, \textit{mass murder (target)}, \textit{Andrey Shpagonoy (attacker)}, and \textit{fireamrs (instrument)}.  

One key challenge in implicit EAE is how to capture the long-range dependency between arguments and the given trigger. Previous methods mainly rely on the pair-wise learning paradigm that models the direct relation between each argument and trigger. \citet{ebner-etal-2020-RAMS} determined the best argument span according to the matching score between the candidate span and the trigger-role representations. \citet{du2020event,li2020event,liu-etal-2020-event} devised the trigger and role-specific questions to locate the argument spans, by treating it as Machine Reading Comprehension (MRC). Recent approaches tend to incorporate longer dependency to enhance argument extraction. \citet{du2020document,du2020document1} introduced a generative transformer-based framework, which encoded the document-level context and decoded arguments one-by-one. Whereas, the order of argument generation is difficult to determine, which is vulnerable to the error propagation as shown in previous studies \cite{xiangyu-etal-2021-capturing}.


Another challenge is how to obtain good performance in low-data scenarios. Due to the long-range complex dependency, it is usually expensive to obtain large-scale  labeled data for training. Even the performance of the state-of-the-art models drops sharply when the argument is two-sentence away from the trigger \cite{zhang-etal-2020-twosteps}. To alleviate this problem, \citet{liu-etal-2021-DocMRC} proposed implicit and explicit data augmentation methods based on MRC. Whereas, these methods rely on the quality of external data, where the domain shift or noise will have side effects on the performance. Recently, prompt-based learning has been investigated to elicit knowledge from pre-trained language models (PLMs) for low-data scenarios \cite{prefix_prompt,2021softprompt}. \citet{li-etal-2021-bart-gen} incorporated a template with masked arguments into the encoder-decoder framework, which conditionally generated the complete template by argument filling, and achieved good performance in the zero-shot scenarios. Whereas, the template is defined based on the event ontology that is not available in many cases. Moreover, the template text description can not well reflect the complex dependency between multiple arguments and triggers.
Thus, more effective prompt templates for modeling the complex dependencies remain to be studied.


To resolve the above two challenges, we propose a \textbf{Cu}rriculum learning based \textbf{P}rompt tuning (CUP) approach for implicit EAE in this paper. Specifically, we first parse the document as an Abstract Meaning Representation (AMR) graph \cite{banarescu2013abstract}, which can well capture the long-range dependency by an explicit path from arguments to the given trigger. To obtain an effective order for argument extraction, we present a curriculum learning framework that learns to extract arguments by four stages with increasing difficulties, inspired by the cognitive process of learning from easy-level to hard-level gradually. The four learning stages are \textit{Center-based Extraction}, \textit{Neighborhood-based Extraction}, \textit{Document-based Extraction}, and \textit{Document-based Extraction without Clues}. The learning difficulty is defined by the hops between the extraction part and the trigger node in the AMR graph. 

In each stage, we utilize a prompt-based encoder-decoder model, which alleviates the data insufficiency problem by eliciting related knowledge from PLMs. It is notable that the prompt templates are adapted according to the learning difficulties in different stages. For the first two stages that the sentences are near the trigger nodes (easy-level), we utilize the textual templates from annotation guidelines. While for the last two stages that focus on document extraction (hard-level), we additionally devise various graph-based templates, which enhance the reasoning ability across multiple arguments and triggers. For inference in testing, we present a pipeline decoding method, which decodes the arguments gradually from the trigger around sentences to the whole document.



Experiments are performed on two well-known benchmark datasets, namely RAMS\cite{ebner-etal-2020-RAMS} and WikiEvents\cite{li-etal-2021-bart-gen}, and the results show the effectiveness of our proposed approach. Particularly, we outperform the state-of-the-art approaches in fully-supervised and low-data scenarios. The main contributions of our work are as follows:
\begin{itemize}
    \item To the best of our knowledge, it is the first attempt to perform implicit EAE with four explicit curriculum learning stages, which well capture the long-range dependency and extraction order.
    \item Different templates are adapted for prompt tuning in different learning stages, which can further enhance the reasoning ability of the target argument in low-data scenarios.
    \item We conduct elaborate analyses of the experimental results on two benchmark datasets, and provide a better understanding of the effectiveness of our approach.
\end{itemize}

\begin{figure*}[t]
    \centering
    \includegraphics[width=\linewidth]{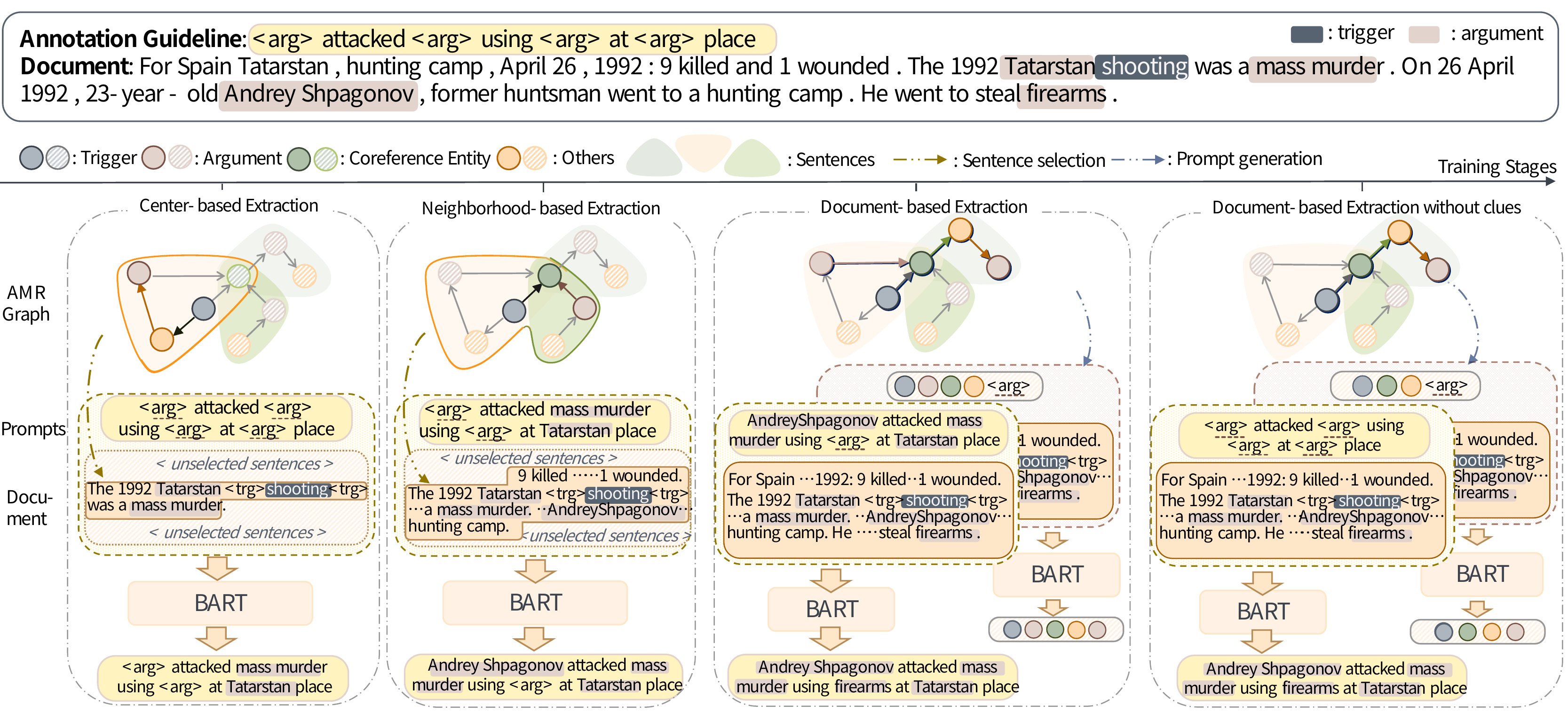}
    \caption{ Overview of curriculum learning with prompts during training. The training stages go on from left to right. In the first two stages, we select sentences containing or surrounding the trigger word. The prompts are guideline-based, gradually filled with clues. In the last two stages, the input are expanded to the full document, with additional structuralized prompts generated from the AMR graph.  }
    \label{fig:modelOverview}
\end{figure*}

\begin{figure}[h]
    \centering
    \includegraphics[width=\linewidth]{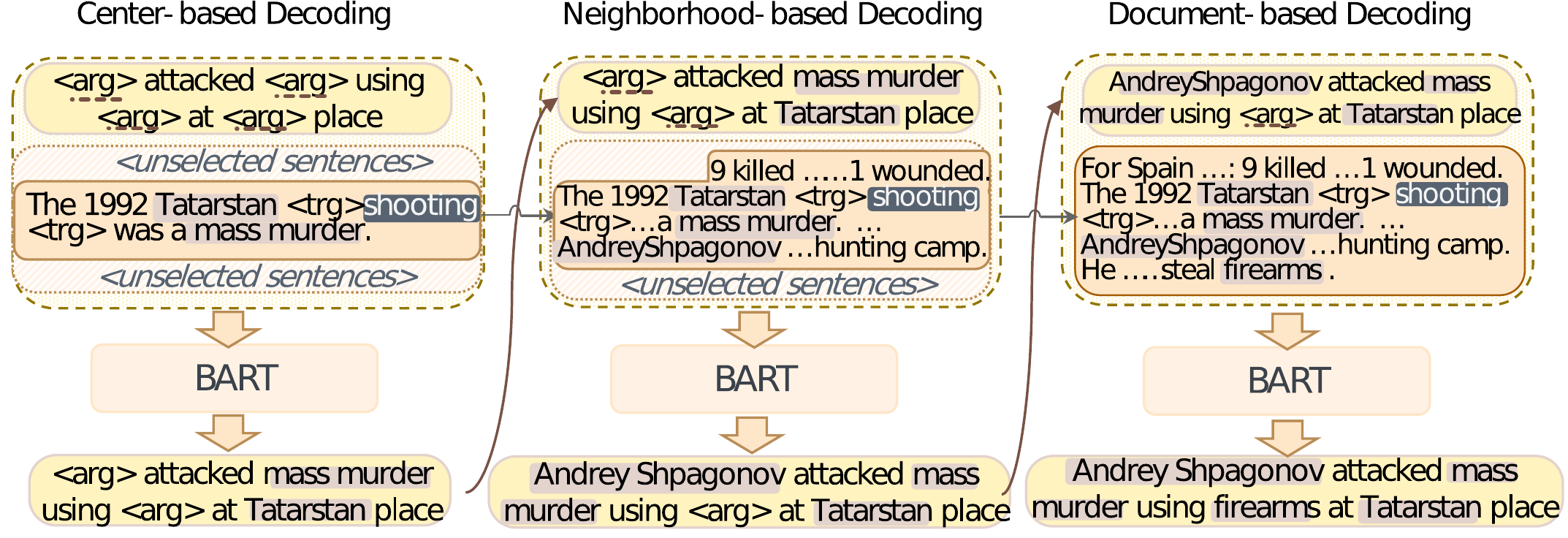}
    \caption{ Pipeline decoding during inference.}
    \label{fig:decoder}
\end{figure}
\section{Our Approach}
The framework of our proposed approach is illustrated in Figure \ref{fig:modelOverview}. Specifically, we first construct an AMR graph for the document that describes the event, which well reflects the long-range dependency between arguments and the given trigger. Then, different parts of the document are selected according to the distance with the trigger node in the AMR graph, which are then used for implicit EAE by four-staged curriculum learning. In particular, we integrate different prompt templates into the pre-trained language model (BART) \cite{lewis-etal-2020-bart} in different stages, which generates the complete template by argument prediction. During inference, we utilize a pipeline decoder, which also generates the arguments gradually according to the learning progress as shown in Figure \ref{fig:decoder}.


\subsection{AMR Graph Construction}
Abstract Meaning Representation (AMR) graph \cite{banarescu2013abstract} is a sembanking language that represents the semantic structure of a sentence. Compared to traditional dependency parsing, the nodes in AMR are entities instead of words, and the relations are more logical and less vulnerable to the syntactic representation or word order variations. Therefore, we utilize AMR to model the interdependency between arguments and the given trigger. 
First, we utilize a transition-based AMR parser \cite{zhou-etal-2021-amr-parser} trained on AMR 2.0 annotations to generate an AMR graph for each sentence in a document. Considering that some entities will be split into multiple tokens and represented by several nodes, we merge them into a single node to obtain a complete representation. Then, we link sentence-level AMR graphs together by coreferenced entities shared in the document. Here we employ ready-made coreference resolver provided by AllenNLP\footnote{https://demo.allennlp.org/coreference-resolution} \cite{Lee2018HigherorderCR}. Finally, a large AMR graph corresponding to the document is obtained. Formally, give a document $D$, its document-level AMR graph is represented as $G^D=(V^D,E^D)$, where $V^D$ represents the nodes including arguments and the given trigger, and $E^D$ indicates the edges that reflect the dependency between nodes. 

\subsection{Curriculum Learning with Prompts}


Motivated by the cognitive progress of humans and animals that learn knowledge from easy to hard gradually, we propose a curriculum learning \cite{curriculum-learning} framework to resolve implicit EAE by several stages, which learns to extract arguments from trigger around sentences to the whole document. Specifically, the learning stages are defined according to the number of hops from the given trigger to a target argument in the document AMR graph, which well reflects the dependence length and extraction difficulties. Four stages are explicitly presented, namely Center-based Extraction (Cent Ex), Neighborhood-based Extraction (Neigh Ex), Document-based Extraction (Doc Ex), and Document-based Extraction without Clues (Doc Ex (w/o Cl)). Moreover, we incorporate different prompt-tuning strategies in different learning stages, which can help elicit knowledge from PLMs to boost performance in low-data scenarios. The details of each learning stage are presented as follows.



\noindent \textbf{Center-based Extraction.} Intuitively, the sentence that contains the trigger node is more pertinent to the event, which can be deemed as the centered sentence. In the first stage, we focus on extracting arguments from the centered sentence (i.e., the sentence with orange background in Figure \ref{fig:modelOverview}). In other words, we only need to conduct sentence-level EAE and extract the nearest arguments with the shortest dependency on the trigger in this stage. A pre-trained language model, namely BART \cite{lewis-etal-2020-bart}, is used for argument extraction. Specifically, for a centered sentence $x_i$ with trigger inside, we first use special token `$\langle$ trg $\rangle$' to markup the position of the trigger word and get $\overline{x}_i$ as: 
\begin{equation}
    \overline{x}_i=[t_0,...,t_{trg-1} ,\langle trg \rangle, t_{trg} ,\langle trg \rangle,t_{trg+1}...,t_l]
\end{equation}
where $t_i$ indicates the $i$th token in the sentence, and $trg$ is the index of the trigger word.
Then, $\overline{x}_i$ is wrapped along with a prompt template $p$ by a function $\lambda \left ( \cdot, \cdot  \right ) $ :
\begin{equation}
    X_{p} = \lambda \left ( p,\overline{x}_i \right ) = \langle s \rangle \ p \ \langle / s \rangle  \ \overline{x}_i \  \langle / s \rangle 
\end{equation}
Here, $p$ is manually adapted from event annotation guidelines where all arguments are replaced by a placeholder token $\langle arg \rangle$, $\langle s \rangle$ and $\langle / s \rangle$ indicate the start and end of a sentence in BART respectively. After that, we obtain the input of the encoder, i.e., $X_{p}$, which is then encoded to generate a filled template that replaces the placeholder tokens by the extracted arguments.

\noindent \textbf{Neighborhood-based Extraction.} After extraction from the centered sentence, we expand the neighborhood sentences in which the concept nodes are near the trigger and extracted arguments in the AMR graph. As shown in Figure \ref{fig:modelOverview}, the linkages from arguments in the neighborhood sentences (with green background) to the trigger node go through at least one coreference node, which increases the extraction difficulty. To facilitate extraction, we adapt the manual prompt template by filling placeholders with previously extracted arguments as clues, which is denoted as $p^{c}$. Hence, our model can 
capture relations among extracted arguments and infer cross-sentence arguments more efficiently. Formally, the input of the encoder is represented as: 
\begin{equation}
    X_{p^{c}} = \lambda \left ( p^{c},[x_{i-1};\overline{x}_i;x_{i+1}] \right )
\end{equation}
where $x_{i-1}$ and $x_{i+1}$ represent the neighborhood sentences, and `;' denotes concatenation.


\noindent \textbf{Document-based Extraction.}
In this stage, we expand the extraction range to the whole document. Considering the arguments could span over multiple sentences, we devise a graph-based structuralized prompt in addition to the above non-structuralized textual prompt, which aims to enhance the reasoning ability for the arguments that have long-range dependency on the trigger. To be specific, we first find the smallest subgraph of $G^D$ that contains the trigger ($trg$), the nearest argument in the centered sentence ($a_c$), and the target argument to be extracted ($a_t$). Then, the subgraph is transformed into a sequence by a linearization function to form the prompt template, which is formulated as:
\begin{equation}
    p_{g}^c= Linear(subgraph(G^D,trg,a_c,a_t))
\end{equation}
where $subgraph$ is a function that locates the smallest subgraph from $G^D$ which contains the nodes corresponding to $trg$, $a_c$ and $a_t$, $Linear$ is a function that transforms the subgraph into a sequence, which can be implemented by the graph traversing algorithm as demonstrated in \cite{2021SPRINGtoRule}. Here, we apply the depth-first searching (DFS) algorithm to obtain the sequence.


With the graph-based prompt $p_{g}^c$ and previous text-based prompt $p^{c}$ (with extracted clues), we combine them with the document by the function $\lambda \left ( \cdot, \cdot  \right ) $ to obtain two prompt-based representations as the input of the encoder respectively:
\begin{equation}
\begin{split}
    X_{p^c} = \lambda \left ( p^{c},\overline{D} \right )\ \
    X_{p_{g}^c}= \lambda \left ( p_{g}^c,\overline{D} \right )
\end{split}
\end{equation}
where $\overline{D}$ is adapted from $D$ by surrounding the trigger with special `$\langle trg \rangle$' tokens.

\noindent \textbf{Document-based Extraction without Clues.}
The last stage is document extraction without using previously extracted arguments as clues, which is the hardest among all stages. In addition to the raw manual prompt used in the first stage (i.e., $p$), we devise an additional structuralized prompt without any previous clues. Specifically, the shortest path between the trigger node and target argument node in the AMR graph is used to form the structuralized prompt without clues:
\begin{equation}
    p_{g}= path(G^D,trg,a_t)
\end{equation}
where $path$ is a function to find the shortest path from the trigger $trg$ to the target argument $a_t$ in the AMR graph $G^D$. 

In this stage, we also have two varieties of inputs for the encoder, which aims to enhance document-based argument extraction with original textual prompt and structuralized path based prompt:
\begin{equation}
\begin{split}
    X_{p} = \lambda \left ( p,\overline{D} \right ) \ \ 
    X_{p_{g}}= \lambda \left ( p_{g},\overline{D} \right )
\end{split}
\end{equation}

\subsection{Training Objective}
For each stage, we feed the related text with prompt, denoted as $X$, into the BART encoder, and obtain the conditional generation probability for each output token $o_c$ by decoder as:
\begin{equation}
    p(o_c|o_{<c},X)=\rm{BART}(X)
\end{equation}
The model parameters are optimized by minimizing the negative loglikelihood over the prompt with gold arguments:
\begin{equation}
     \mathcal{L}(X) = -\sum_{c=1}^C \log(p(o_c|o_{<c},X))
\end{equation}
where $C$ denotes the length of the prompt template.

For the last two training stages, since there are two varieties of input with different prompts, we calculate the decoding loss for each input separately, and introduce a hyperparameter $\alpha$ to combine them as follows:
\begin{equation}
    \mathcal{L} = \mathcal{L}(X_{p}) + \alpha \mathcal{L}(X_{p_{g}})
    \label{loss_all}
\end{equation}

\subsection{Pipeline Decoding}
Most previous methods decode arguments according to their orders in the document or guideline \cite{du2020document1,li-etal-2021-bart-gen}, which may violate the cognitive process of learning from easy-level to hard-level \cite{curriculum-learning}. In contrast, we present a pipeline decoding for implicit EAE during inference, which is consistent with the learning process in the training period. Specifically, our pipeline decoding consists of three stages as shown in Figure \ref{fig:decoder}, namely Center-based Decoding (Cent Dec), Neighborhood-based Decoding (Neigh Dec), and Document-based Decoding (Doc Dec), which decode the prompt template by filling placeholders with arguments from the centered sentence, neighborhood sentences and full document gradually. Notably, only the textual prompts based on guidelines are used in all stages since the graph-based prompts are unavailable due to the unlabeled target argument in the test data. All previous extracted arguments are used as clues to boost further decoding.

\section{Experimental Setup}
\textbf{Datasets.} We conduct extensive experiments on two implicit EAE datasets, namely RAMS \cite{ebner-etal-2020-RAMS} and WikiEvents \cite{li-etal-2021-bart-gen}, which have been widely used in previous studies \cite{li-etal-2021-bart-gen,liu-etal-2021-DocMRC}. The RAMS dataset includes 3,993 paragraphs, annotated with 139 event types and 65 semantic roles. In the WikiEvents dataset, 246 documents are collected and annotated with 50 event types and 59 semantic roles. Our experiments are conducted under informative argument setting\footnote{ In this setting, the most informative mention of the argument is treated as the ground truth, which is more challenging.}. We follow the official data split of each dataset. Detailed statistics are listed in Table \ref{tab:data stat}. The performance is evaluated by Precision (P), Recall (R), and F1 score (F1), based on the Exact Match (EM) criterion: only when the prediction matches the gold one exactly, it is deemed as a correct prediction. 

\noindent \textbf{Implementations.}
Considering that most previous studies are based on the base-scale models (e.g., BERT-base), we employ a vanilla BART-base model in our approach for fairness of comparison. To adapt to the input length of the BART model, we truncate the documents in WikiEvents that are much longer than those in RAMS, and select the trigger surrounding sections with at most nine sentences for experiments. The experiments are conducted on a single Nividia GeForce RTX 3090. For the four stages of curriculum learning, the learning rate decreases as the learning process goes on, which is set as 1e-4, 5e-5, 3e-5 and 2e-5 respectively. The best batch size is searched from the set of \{16, 32\} for each stage. The hyper-parameter of $\alpha$ that balances the two losses in Formula (\ref{loss_all}) ranges from 0 to 1 with a step of 0.1, and we set it to 0.7 as a reliable setting. All other parameters are optimized by the Adam \cite{kingma2014adam} algorithm. The optimal parameters are obtained based on the development set, then used for evaluation on test set.\footnote{Our code is available at \url{https://github.com/linmou/CUP}} 

\begin{table}[!t]
\centering
\begin{tabular}{ccccc}
\hline
DataSet                & Split & Doc  & Event & Argument \\ \hline
\multirow{3}{*}{RAMS}  & Train & 3194 & 7329  & 17026    \\
                       & Dev   & 399  & 924   & 2188     \\
                       & Test  & 400  & 871   & 2023     \\ \hline
\multirow{3}{*}{WikiEvents} & Train & 206  & 3241  & 4542     \\
                       & Dev   & 20   & 345   & 428      \\
                       & Test  & 20   & 365   & 556      \\ \hline
\end{tabular}
\caption{Data statistics of RAMS and WikiEvents}
\label{tab:data stat}
\end{table}

\noindent  \textbf{Baselines.}
We compare our approach with the recent advanced baselines: 1) SpanSel \cite{ebner-etal-2020-RAMS} is a method based on span ranking, which enumerates each possible span in a document to identify the argument; 2) Head-Expand \cite{li-etal-2019-entity} extends SpanSel by first identifying an argument's head and then its region; 3) BART-Gen \cite{ebner-etal-2020-RAMS} treats implicit EAE as a generation task, and generates the arguments based on BART architecture\footnote{Here we define BART-Gen as its BART-base reproduction for a fair comparison \cite{liu-etal-2021-DocMRC}.}; 4) DocMRC \cite{liu-etal-2021-DocMRC} models implicit EAE as a MRC problem, which devises the role-specific questions for argument extraction; 5) DocMRC (IDA) is an enhanced version of DocMRC, which uses implicit data augmentation by pre-training on other related tasks, including question answering, natural language inference and sentence level event argument extraction.

\begin{table}[!t]
\small
\begin{tabular}{l|ccc|ccc}
\hline
\multirow{2}{*}{Method}              & \multicolumn{3}{c|}{RAMS}                     & \multicolumn{3}{c}{WikiEvents} \\ 
        & P             & R             & F1            & P        & R        & F1       \\ \hline
SpanSel       & 38.2          & 43.6          & 40.7          & 57.8     & 29.2     & 38.3     \\
Head-Expand & -             & -             & 41.8          & 57.8     & 30.6     & 40.0     \\
BART-Gen      & 41.9          & 42.5          & 42.2          & 60.0     & 32.0     & 41.7     \\
DocMRC        & 42.6          & 46.1          & 44.3          & \textbf{61.7}     & 32.0     & 42.1     \\ 
DocMRC (IDA)  & 43.4          & \textbf{48.3} & 45.7          & 60.2     & 33.7     & 43.3     \\ \hline
CUP (Ours)    & \textbf{46.0} & 47.0          & \textbf{46.5} & 49.0     & \textbf{45.8}  & \textbf{47.4}          \\ \hline
\end{tabular}
\caption{Results on RAMS and WikiEvents.}
\label{tab:main results}
\end{table}

\section{Results and Analyses}
\subsection{Overall Performance}
Table \ref{tab:main results} shows the performance of our approach and the state-of-the-art baselines on RAMS and WikiEvents. For fairness of comparison, all the results are under the settings that the event types are known. We have the following observations. \textbf{First}, we achieve the best performance on the two datasets in most cases, indicating the superiority of our proposed approach without using any external data for pre-training or labelling.
\textbf{Second}, our approach that integrates BART in our curriculum learning framework surpasses the BART-GEN baseline with a large magnitude, indicating that we can make the most potential of the BART architecture by curriculum learning based prompt tuning. 
\textbf{Third}, compared with the best baseline DocMRC (IDA) that relies on external data for augmentation, our approach outperforms it with an improvement up to 9.47\% in terms of F1 on WikiEvents, which indicates the effectiveness of our prompt tuning strategies in alleviating the data insufficiency problem.

\subsection{Ablation Studies}
We conduct ablation studies to investigate the effectiveness of each component of our approach, and the results are shown in Table \ref{tab:ablations}. First, we remove one of the four curriculum learning stages each time, noted as - $\rm{Cent \ Ex}$, - $\rm{Neigh \ Ex}$, - $\rm{Doc \ Ex}$, - $\rm{Doc \ Ex \ (w/o \ Cl)}$. Then, we verify the effectiveness of graph-based prompt, illustrated by - $\rm{p_{g}}$. For pipeline decoding, we only remove Center-based Decoding (- $\rm{Cent \ Dec}$) and Neighborhood-based Decoding (- $\rm{Neigh \ Dec}$) respectively, since Document-based Decoding can not be removed for complete argument extraction. 

We can observe that each curriculum learning stage can help boost the performance of implicit EAE. To be specific, without Center-based Extraction, the performance on both RAMS and WikiEvents drops dramatically by about 2 points in terms of F1. This justifies that argument extraction from the centered sentence that contains the trigger is indispensable. Similar observations can be found in other stages, indicating the effectiveness of each stage for enhancing argument extraction. In particular, the last stage ($\rm{Doc \ Ex \ (w/o \ Cl)}$) also plays an important role as the third stage ($\rm{Doc \ Ex}$), though the only difference lies in that it does not use the extracted arguments as clues for extraction. This finding verifies the necessity of using the last stage as the hardest learning task to boost implicit EAE. Regarding prompt tuning, it is observed the graph-based prompt $p^{g}$ brings an improvement up to 6.04\% on WikiEvents with respect to F1, demonstrating the superiority of graph-based prompt for argument reasoning over long-range dependency. For pipeline decoding, we have similar observations. Each decoding step is effective, and decoding arguments from the centered sentence ($\rm{Cent \ Dec}$) is also more important than other steps. 



\begin{table}[!t]
\centering
\small
\begin{tabular}{l|ccc|ccc}
\hline
\multirow{2}{*}{Model} & \multicolumn{3}{c|}{RAMS}                     & \multicolumn{3}{c}{WikiEvents}                \\
                       & P             & R             & F1            & P             & R             & F1            \\ \hline
CUP                    & 46.0          & \textbf{47.0} & \textbf{46.5} & 49.0          & \textbf{45.8} & \textbf{47.4} \\ \hline
- Cent Ex              & 43.1          & 45.5          & 44.2          & 47.6          & 43.5          & 45.4          \\
- Neigh Ex             & 44.8          & 46.1          & 45.5          & 49.6          & 43.5          & 46.3          \\
- Doc Ex               & 46.2          & 44.3          & 45.2          & 48.1          & 41.5          & 44.5          \\
- Doc Ex (w/o Cl)      & 42.7          & 46.9          & 44.7          & 49.0          & 42.7          & 45.6          \\
- p$_{\rm{g}}$         & 45.3          & 45.6          & 45.5          & 46.6          & 42.9          & 44.7          \\ \hline
- Cent Dec             & 43.5          & 44.9          & 44.2          & 46.8          & 44.9          & 45.8          \\
- Neigh Dec            & \textbf{46.9} & 44.7          & 45.8          & \textbf{50.2} & 43.5          & 46.6          \\ \hline
\end{tabular}
\caption{ Results of ablation studies.}
\label{tab:ablations}
\end{table}

	

\subsection{Impact of Trigger-Argument Distance} 
To investigate whether our approach can well capture the long-range dependency between the trigger and arguments, we use the sentence-level distance to roughly model the dependency. Each sample in RAMS contains five sentences with explicit sentence boundaries. While for WikiEvents, in many cases, it splits sentences inaccurately. Thus, we split the document into sentences with spaCy\footnote{https://spacy.io/}, and select a five-sentence paragraph around the trigger for experiments. Table \ref{tab:impact of dis} illustrates the impact of sentence-level distance from triggers to arguments. We compare with two state-of-the-art baselines shown in Table \ref{tab:main results}. Note that DocMRC (IDA) additionally uses external data for training, which is excluded here for a fair comparison. We observe that our approach outperforms baselines in most cases with various distances. It is also interesting to find that the performance of all the models does not always drop when the distance increases, especially for the WikiEvents dataset. This may be related to the various syntax and discourse structures in event descriptions, and the sentence-level distance can not completely reflect the complex dependency between arguments and the trigger. Overall, our approach performs better than the baselines even when the trigger-argument distance increases to two sentences (Table \ref{tab:impact of dis}) or the whole document (Table \ref{tab:main results}).



\begin{table}[h]
    \begin{subtable}
        \centering
        \begin{tabular}{lccccc}
\hline
\multirow{2}{*}{Method}         & \multicolumn{5}{c}{Trigger-Argument Distance on RAMS}                        \\ 
      & -2$_{[4\%]}$             & -1$_{[8\%]}$         & 0$_{[83\%]}$         & 1$_{[4\%]}$     & 2$_{[2\%]}$    \\ \hline
BART-Gen\SP{*}    & 17.7              & 16.8          & 44.8              & 16.6              & \textbf{9.0}           \\
DocMRC      & 15.2              & 12.6          & 48.6              & \textbf{20.5}     & 3.4          \\
CUP (Ours)  & \textbf{19.3}  & \textbf{22.2} & \textbf{49.6}     &\underline{17.4}  & \underline{8.6} \\ \hline
\end{tabular}
\end{subtable}
\begin{subtable}
\centering
\begin{tabular}{lccccc}
\hline
\multirow{2}{*}{Method}            & \multicolumn{5}{c}{Trigger-Argument Distance on WikiEvents}                        \\ 
      & -2$_{[8\%]}$          & -1$_{[10\%]}$     & 0$_{[67\%]}$        & 1$_{[8\%]}$         & 2$_{[7\%]}$             \\ \hline
BART-Gen    & 11.1           & 6.9           & 43.2              & 2.9                & 7.7        \\
DocMRC      & 13.9           & 5.6           & 44.4             & \textbf{7.3}        & 8.0          \\
CUP (Ours)  & \textbf{15.7}           & \textbf{7.7}  & \textbf{51.4}     &  \underline{6.7}             & \textbf{8.9}        \\ \hline
\end{tabular}
    \end{subtable}
     \caption{Impact of the trigger-argument distance. The results marked with $*$ is from
     \protect\cite{liu-etal-2021-DocMRC}. 
     Other results are re-produced due to the different experimental settings, such as sentence split method or training data. }
     \label{tab:impact of dis}
\end{table}

\subsection{Performance in Low-data Scenarios}
\begin{figure}[t]
\tiny
\centering
\includegraphics[width=0.49\columnwidth]{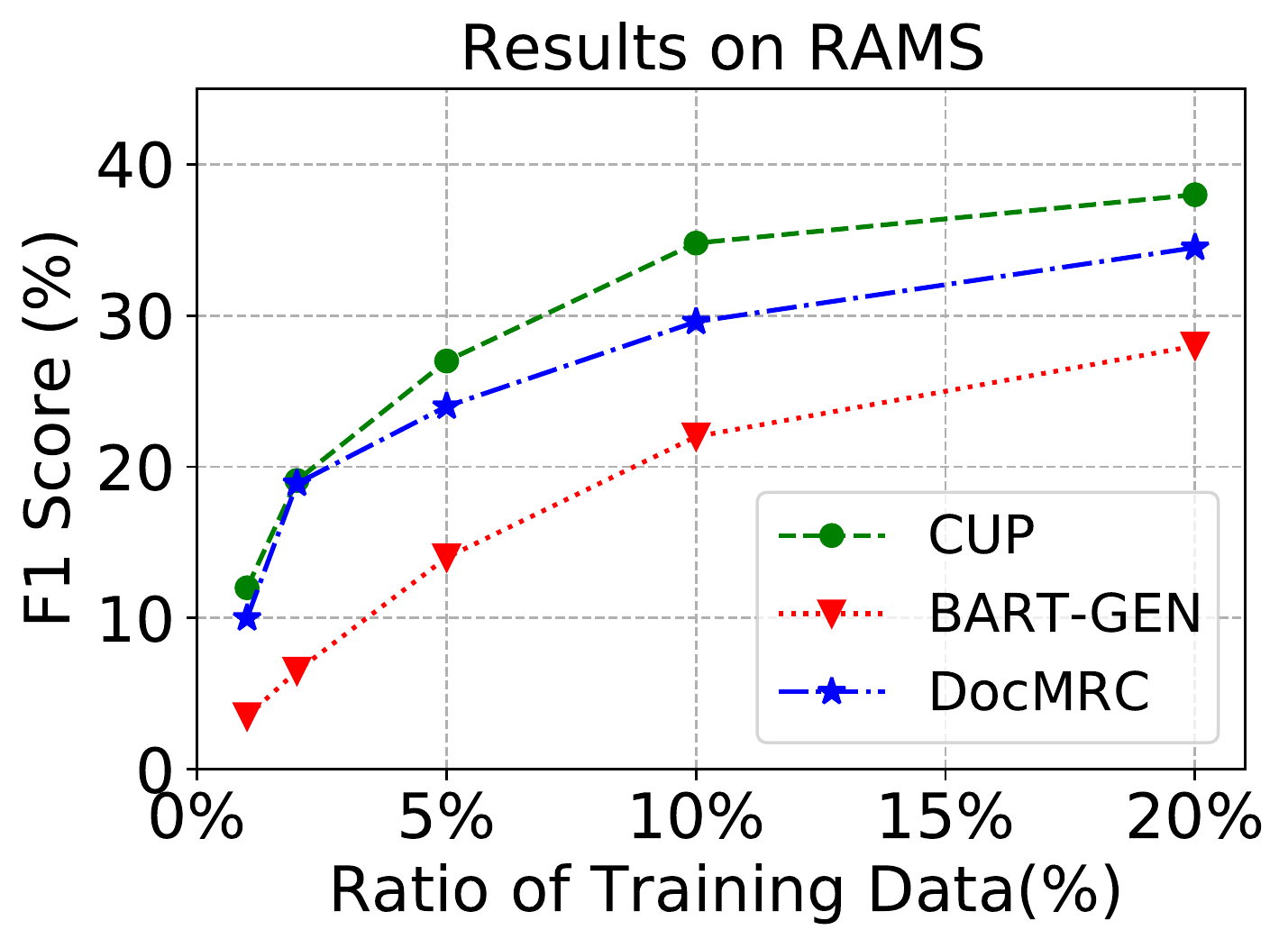}
\includegraphics[width=0.49\columnwidth]{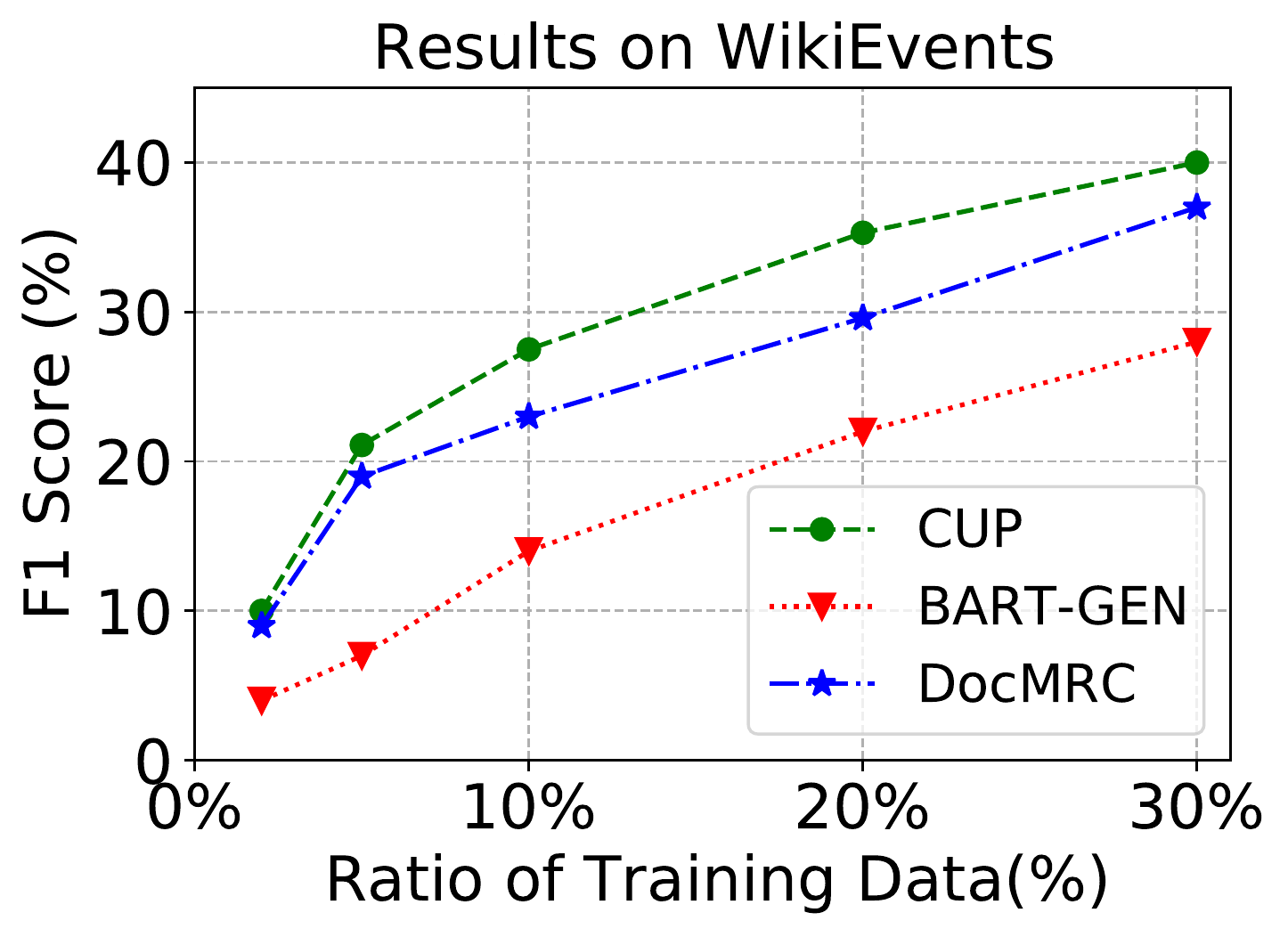}
\caption{ Results in the data-low scenario on RAMS and WikiEvents} 
	\label{fig:low resource}
\end{figure}

To examine the performance under low-data scenarios, we vary the ratio of the training data from 1\% to 30\%, and the results are shown in Figure \ref{fig:low resource}. It is observed that our approach outperforms the two state-of-the-art baselines on both datasets. In particular, our approach overtakes DocMRC trained by 20\% of the training data, with only half of the samples (10\%) on RAMS dataset. All these observations verify the effectiveness of our prompt tuning strategies that elicit related knowledge from pre-trained language models, which can help alleviate the data insufficiency problem.



\section{Related Work}
\textbf{Event Argument Extraction (EAE)} aims to identify arguments from text with a given trigger. Sentence-level EAE has achieved great progress in previous studies \cite{chen-etal-2015-event,liu-etal-2018-jointly}. While implicit EAE that arguments could span over multiple sentences is not well studied. Previous works treat it as a semantic role labeling task \cite{zhang-etal-2020-twosteps}. Recently, there is a trend to formulate implicit EAE as Machine Reading Comprehension (MRC) \cite{du2020event}, where manual questions are designed to locate the answer as the target argument. Whereas, the MRC-based methods focus on learning the pair-wise trigger-argument relations, which neglect the long-range dependency between trigger and arguments. In addition, it is difficult to obtain the training data due to the expensive cost. \cite{liu-etal-2021-DocMRC} attempted to alleviate the data sparsity problem by implicit and explicit data augmentations. However, these methods rely on external data, which is vulnerable to domain shift or noise. 

\noindent \textbf{Prompt Tuning} is introduced as a new paradigm of natural language processing \cite{prompt_survey}. 
Rather than introducing a classifier during fine-tuning, prompt-based methods wrap input context with a continuous or discrete vector with unfilled slots called prompt.
Specifically, prompt engineering is a key step for prompt tuning, which creates a proper prompt template for the task. \citet{LMasKB} first proposed a manual designed prompt which transformed classification as a cloze filling task. Other studies attempt to explore automatic searched prompts to alleviate the manual cost \cite{prefix_prompt}\cite{2021softprompt}. 
Due to the complexity of implicit EAE, the appropriate prompt tuning methods remain to be studied. In this work, we explore different prompt tuning strategies in different learning stages.


\noindent \textbf{Curriculum Learning} is a learning tactic proposed by \citet{curriculum-learning}, where a model is trained by training data with increasing complexity. \citet{wei-etal-2021-triggerNotSufficient} introduced a curriculum knowledge distillation strategy to incorporate related arguments for implicit EAE. Whereas, the learning difficulty is not explicitly defined, and the argument relations are not included in inference. In this paper, we propose a well-defined multi-staged curriculum learning framework, which models the interdependency for both training and inference.


\section{Conclusions}
In this paper, we propose a curriculum learning based prompt tuning approach for implicit EAE. Extensive experiments have verified the effectiveness of each learning stage. Furthermore, our approach consistently outperforms the state-of-the-art models in both full-training or low-data scenarios, by incorporating different prompt-tuning strategies in different learning stages. In particular, our graph-based prompt significantly boosts the performance by up to 6.04\% in terms of F1. In the future, we will explore more prompts to enhance reasoning for target arguments with long-range dependency. 

\section*{Acknowledgments}
This research is funded by the National Key Research and Development Program of China (No. 2021ZD0114002), the National Nature Science Foundation of China (No. 61906045), and the Science and Technology Commission of Shanghai Municipality Grant (No. 21511100402).

\small
\bibliographystyle{named}
\bibliography{ijcai22}

\begin{thebibliography}{}

\bibitem[\protect\citeauthoryear{Banarescu \bgroup \em et al.\egroup
  }{2013}]{banarescu2013abstract}
Laura Banarescu, Claire Bonial, Shu Cai, Madalina Georgescu, Kira Griffitt, Ulf
  Hermjakob, Kevin Knight, Philipp Koehn, Martha Palmer, and Nathan Schneider.
\newblock Abstract meaning representation for sembanking.
\newblock In {\em ACL Workshop}, pages 178--186, 2013.

\bibitem[\protect\citeauthoryear{Bengio \bgroup \em et al.\egroup
  }{2009}]{curriculum-learning}
Yoshua Bengio, J\'{e}r\^{o}me Louradour, Ronan Collobert, and Jason Weston.
\newblock Curriculum learning.
\newblock In {\em ICML}, ICML '09, page 41–48, New York, NY, USA, 2009.
  Association for Computing Machinery.

\bibitem[\protect\citeauthoryear{Bevilacqua \bgroup \em et al.\egroup
  }{2021}]{2021SPRINGtoRule}
Michele Bevilacqua, Rexhina Blloshmi, and Roberto Navigli.
\newblock One spring to rule them both: Symmetric amr semantic parsing and
  generation without a complex pipeline.
\newblock In {\em AAAI}, 2021.

\bibitem[\protect\citeauthoryear{Chen \bgroup \em et al.\egroup
  }{2015}]{chen-etal-2015-event}
Yubo Chen, Liheng Xu, Kang Liu, Daojian Zeng, and Jun Zhao.
\newblock Event extraction via dynamic multi-pooling convolutional neural
  networks.
\newblock In {\em IJCNLP}, pages 167--176, Beijing, China, July 2015.

\bibitem[\protect\citeauthoryear{Chen \bgroup \em et al.\egroup
  }{2020}]{chen-etal-2020-joint-modeling}
Yunmo Chen, Tongfei Chen, and Benjamin Van~Durme.
\newblock Joint modeling of arguments for event understanding.
\newblock In {\em Proceedings of the First Workshop on Computational Approaches
  to Discourse}, pages 96--101, Online, November 2020. Association for
  Computational Linguistics.

\bibitem[\protect\citeauthoryear{Du and Cardie}{2020a}]{du2020document}
Xinya Du and Claire Cardie.
\newblock Document-level event role filler extraction using multi-granularity
  contextualized encoding.
\newblock In {\em ACL}, pages 8010--8020, 2020.

\bibitem[\protect\citeauthoryear{Du and Cardie}{2020b}]{du2020event}
Xinya Du and Claire Cardie.
\newblock Event extraction by answering (almost) natural questions.
\newblock In {\em EMNLP}, pages 671--683, 2020.

\bibitem[\protect\citeauthoryear{Du \bgroup \em et al.\egroup
  }{2020}]{du2020document1}
Xinya Du, Alexander Rush, and Claire Cardie.
\newblock Document-level event-based extraction using generative
  template-filling transformers.
\newblock {\em arXiv e-prints}, pages arXiv--2008, 2020.

\bibitem[\protect\citeauthoryear{Ebner \bgroup \em et al.\egroup
  }{2020}]{ebner-etal-2020-RAMS}
Seth Ebner, Patrick Xia, Ryan Culkin, Kyle Rawlins, and Benjamin Van~Durme.
\newblock Multi-sentence argument linking.
\newblock In {\em ACL}, pages 8057--8077, Online, July 2020.

\bibitem[\protect\citeauthoryear{Kingma and Ba}{2014}]{kingma2014adam}
Diederik~P Kingma and Jimmy Ba.
\newblock Adam: A method for stochastic optimization.
\newblock {\em arXiv preprint arXiv:1412.6980}, 2014.

\bibitem[\protect\citeauthoryear{Lee \bgroup \em et al.\egroup
  }{2018}]{Lee2018HigherorderCR}
Kenton Lee, Luheng He, and L.~Zettlemoyer.
\newblock Higher-order coreference resolution with coarse-to-fine inference.
\newblock In {\em NAACL-HLT}, 2018.

\bibitem[\protect\citeauthoryear{Lester \bgroup \em et al.\egroup
  }{2021}]{2021softprompt}
Brian Lester, Rami Al-Rfou, and Noah Constant.
\newblock The power of scale for parameter-efficient prompt tuning.
\newblock In {\em EMNLP}, pages 3045--3059, Online and Punta Cana, Dominican
  Republic, November 2021.

\bibitem[\protect\citeauthoryear{Lewis \bgroup \em et al.\egroup
  }{2020}]{lewis-etal-2020-bart}
Mike Lewis, Yinhan Liu, Naman Goyal, Marjan Ghazvininejad, Abdelrahman Mohamed,
  Omer Levy, Veselin Stoyanov, and Luke Zettlemoyer.
\newblock {BART}: Denoising sequence-to-sequence pre-training for natural
  language generation, translation, and comprehension.
\newblock In {\em ACL}, pages 7871--7880, Online, July 2020.

\bibitem[\protect\citeauthoryear{Li and Liang}{2021}]{prefix_prompt}
Xiang~Lisa Li and Percy Liang.
\newblock Prefix-tuning: Optimizing continuous prompts for generation.
\newblock In {\em ACL-IJCNLP}, pages 4582--4597, Online, August 2021.

\bibitem[\protect\citeauthoryear{Li \bgroup \em et al.\egroup
  }{2019}]{li-etal-2019-entity}
Xiaoya Li, Fan Yin, Zijun Sun, Xiayu Li, Arianna Yuan, Duo Chai, Mingxin Zhou,
  and Jiwei Li.
\newblock Entity-relation extraction as multi-turn question answering.
\newblock In {\em ACL}, pages 1340--1350, Florence, Italy, July 2019.

\bibitem[\protect\citeauthoryear{Li \bgroup \em et al.\egroup
  }{2020}]{li2020event}
Fayuan Li, Weihua Peng, Yuguang Chen, Quan Wang, Lu~Pan, Yajuan Lyu, and Yong
  Zhu.
\newblock Event extraction as multi-turn question answering.
\newblock In {\em EMNLP: Findings}, pages 829--838, 2020.

\bibitem[\protect\citeauthoryear{Li \bgroup \em et al.\egroup
  }{2021}]{li-etal-2021-bart-gen}
Sha Li, Heng Ji, and Jiawei Han.
\newblock Document-level event argument extraction by conditional generation.
\newblock In {\em NAACL}, pages 894--908, 2021.

\bibitem[\protect\citeauthoryear{Liu \bgroup \em et al.\egroup
  }{2018}]{liu-etal-2018-jointly}
Xiao Liu, Zhunchen Luo, and Heyan Huang.
\newblock Jointly multiple events extraction via attention-based graph
  information aggregation.
\newblock In {\em EMNLP}, pages 1247--1256, 2018.

\bibitem[\protect\citeauthoryear{Liu \bgroup \em et al.\egroup
  }{2020}]{liu-etal-2020-event}
Jian Liu, Yubo Chen, Kang Liu, Wei Bi, and Xiaojiang Liu.
\newblock Event extraction as machine reading comprehension.
\newblock In {\em EMNLP}, pages 1641--1651, Online, November 2020.

\bibitem[\protect\citeauthoryear{Liu \bgroup \em et al.\egroup
  }{2021a}]{liu-etal-2021-DocMRC}
Jian Liu, Yufeng Chen, and Jinan Xu.
\newblock Machine reading comprehension as data augmentation: A case study on
  implicit event argument extraction.
\newblock In {\em EMNLP}, pages 2716--2725, Online and Punta Cana, Dominican
  Republic, November 2021.

\bibitem[\protect\citeauthoryear{Liu \bgroup \em et al.\egroup
  }{2021b}]{prompt_survey}
Pengfei Liu, Weizhe Yuan, Jinlan Fu, Zhengbao Jiang, Hiroaki Hayashi, and
  Graham Neubig.
\newblock Pre-train, prompt, and predict: {A} systematic survey of prompting
  methods in natural language processing.
\newblock {\em CoRR}, abs/2107.13586, 2021.

\bibitem[\protect\citeauthoryear{Mostafazadeh~Davani \bgroup \em et al.\egroup
  }{2019}]{mostafazadeh-davani-etal-2019-reporting}
Aida Mostafazadeh~Davani, Leigh Yeh, Mohammad Atari, Brendan Kennedy, Gwenyth
  Portillo~Wightman, Elaine Gonzalez, Natalie Delong, Rhea Bhatia, Arineh
  Mirinjian, Xiang Ren, and Morteza Dehghani.
\newblock Reporting the unreported: Event extraction for analyzing the local
  representation of hate crimes.
\newblock In {\em EMNLP-IJCNLP}, pages 5753--5757, 2019.

\bibitem[\protect\citeauthoryear{Petroni \bgroup \em et al.\egroup
  }{2019}]{LMasKB}
Fabio Petroni, Tim Rockt{\"a}schel, Sebastian Riedel, Patrick Lewis, Anton
  Bakhtin, Yuxiang Wu, and Alexander Miller.
\newblock Language models as knowledge bases?
\newblock In {\em EMNLP-IJCNLP}, pages 2463--2473, Hong Kong, China, November
  2019.

\bibitem[\protect\citeauthoryear{Wei \bgroup \em et al.\egroup
  }{2021}]{wei-etal-2021-triggerNotSufficient}
Kaiwen Wei, Xian Sun, Zequn Zhang, Jingyuan Zhang, Guo Zhi, and Li~Jin.
\newblock Trigger is not sufficient: Exploiting frame-aware knowledge for
  implicit event argument extraction.
\newblock In {\em ACL-IJCNLP}, pages 4672--4682, Online, August 2021.

\bibitem[\protect\citeauthoryear{Xiangyu \bgroup \em et al.\egroup
  }{2021}]{xiangyu-etal-2021-capturing}
Xi~Xiangyu, Wei Ye, Shikun Zhang, Quanxiu Wang, Huixing Jiang, and Wei Wu.
\newblock Capturing event argument interaction via a bi-directional
  entity-level recurrent decoder.
\newblock In {\em IJCNLP}, pages 210--219, Online, August 2021.

\bibitem[\protect\citeauthoryear{Zhan and Jiang}{2019}]{zhan2019survey}
Liying Zhan and Xuping Jiang.
\newblock Survey on event extraction technology in information extraction
  research area.
\newblock In {\em ITNEC}, pages 2121--2126. IEEE, 2019.

\bibitem[\protect\citeauthoryear{Zhang \bgroup \em et al.\egroup
  }{2020}]{zhang-etal-2020-twosteps}
Zhisong Zhang, Xiang Kong, Zhengzhong Liu, Xuezhe Ma, and Eduard Hovy.
\newblock A two-step approach for implicit event argument detection.
\newblock In {\em ACL}, pages 7479--7485, Online, July 2020.

\bibitem[\protect\citeauthoryear{Zhou \bgroup \em et al.\egroup
  }{2021}]{zhou-etal-2021-amr-parser}
Jiawei Zhou, Tahira Naseem, Ram{\'o}n Fernandez~Astudillo, and Radu Florian.
\newblock {AMR} parsing with action-pointer transformer.
\newblock In {\em NAACL}, pages 5585--5598, 2021.

\end{thebibliography}

\end{document}